\documentclass[10pt, a4paper]{article}
\usepackage{lrec}
\usepackage{graphicx}
\usepackage{tabularx}
\usepackage{soul}

\usepackage{makecell}
\usepackage{svg}
\usepackage{float}
\usepackage[utf8]{inputenc}
\usepackage[autostyle]{csquotes}

\usepackage{epstopdf}

\usepackage{hyperref}

\usepackage{CJKutf8}
\usepackage{booktabs,caption,threeparttable}
\usepackage{enumitem}

\newcommand{\korean}[1]{\begin{CJK*}{UTF8}{mj}{#1}\end{CJK*}}

\title{Jejueo Datasets for Machine Translation and Speech Synthesis}

\name{Kyubyong Park, Yo Joong Choe, Jiyeon Ham}

\address{Kakao Brain \\
         20, Pangyoyeok-ro 241, Bundang-gu, Seongnam-si, Gyeonggi-do, Korea \\
         \{kyubyong.park, yj.choe, jiyeon.ham\}@kakaobrain.com \\}

\abstract{
Jejueo  was classified  as  critically  endangered by UNESCO in 2010. Although diverse efforts to revitalize it have been made, there have been few computational approaches. 
Motivated by this, we construct two new Jejueo datasets: \textit{Jejueo Interview Transcripts} (JIT) and \textit{Jejueo Single Speaker Speech} (JSS). 
The JIT dataset is a parallel corpus containing 170k+ Jejueo-Korean sentences, and the JSS dataset consists of 10k high-quality audio files recorded by a native Jejueo speaker and a transcript file. Subsequently, we build neural systems of machine translation and speech synthesis using them. All resources are publicly available via our GitHub repository. We hope that these datasets will attract interest of both language and machine learning communities. \\ \newline  }

\begin{document}

\maketitleabstract

\section{Introduction}
Jejueo, or the Jeju language, is a minority language used on Jeju Island \cite{o2015jejueo}. 
It was classified as critically endangered by UNESCO in 2010.\footnote{\url{http://www.unesco.org/languages-atlas/en/atlasmap.html}}
While there have been many academic efforts to preserve the language \cite{yang2017toward,saltzman-2017-jejueo,yangrevising,yang2018integrating}, data-driven approaches for Jejueo-related language tasks have been rare.

Meanwhile, the natural language processing (NLP) community has observed significant advances in both machine translation \cite{sutskever2014sequence,cho2014learning,vaswani2017attention} and speech synthesis \cite{oord2016wavenet,wang2017tacotron,shen2018natural}, especially driven by deep learning in recent years.

In particular, there has been growing attention towards low-resource scenarios \cite{zoph-etal-2016-transfer,gu-etal-2018-universal}, which pose a unique challenge for existing deep learning methods. 
The challenge is unique not only in the sense that less data is available but also in the sense that low-resource languages often raise different syntactic, morphological, and semantic challenges that are under-explored by systems optimized on major languages such as English, German, French, and Arabic \cite{bender2019benderrule}.
Tasks relevant to Jejueo often involve having to deal with such challenges.

Motivated by the unique challenges that Jejueo presents as well as the lack of Jejueo resources available for computational approaches, we develop a machine-readable Jejueo-Korean parallel corpus and a clean Jejueo single speaker speech dataset.

Our contributions can be summarized as follows:
\begin{itemize}
    \item We present the \textit{Jejueo Interview Scripts} (JIT) dataset, a Jejueo-Korean parallel corpus of more than 170k sentences.
    \item We train neural machine translation models on the JIT dataset so that they can be the baselines for future studies.
    \item We create the \textit{Jejueo Single Speaker Speech} (JSS) dataset of 10k audio files and their transcripts.
    \item We build speech synthesis models with the JSS dataset and examine how various tokenisation strategies affect them.
\end{itemize}

\begin{figure}[t]
    \centering
    \includegraphics[width=8cm,clip=false]{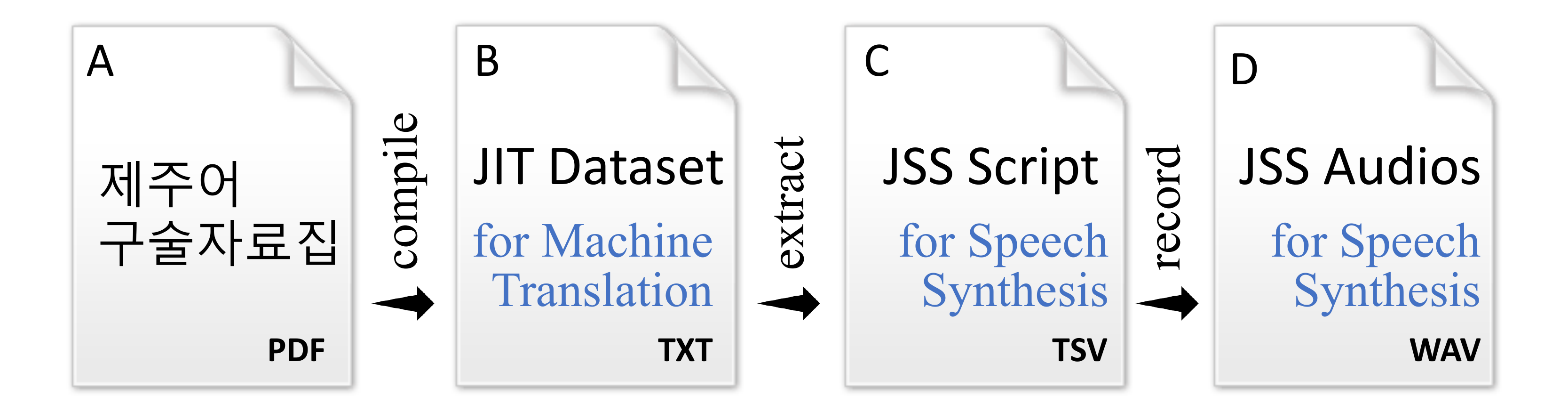}
    \caption{Overview of dataset construction. The original \texttt{pdf} files (A) are compiled into the JIT dataset (B). Part of the Jejueo text in the JIT dataset is extracted and saved as the JSS script file (C). Finally, we ask a native Jejueo speaker to record the script (D).}
    \label{fig:overview}
\end{figure}

The procedure for constructing the datasets is shown in Figure \ref{fig:overview}.
To the best of our knowledge, they are the first publicly available Jejueo datasets for computational tasks, particularly Jejueo-Korean machine translation and Jejueo speech synthesis.

All the resources are released via our GitHub repository\footnote{\url{https://github.com/kakaobrain/jejueo}}.

\section{Jejueo} \label{jejueo}
\begin{figure}
    \centering
    \includegraphics[width=5.5cm,clip=false]{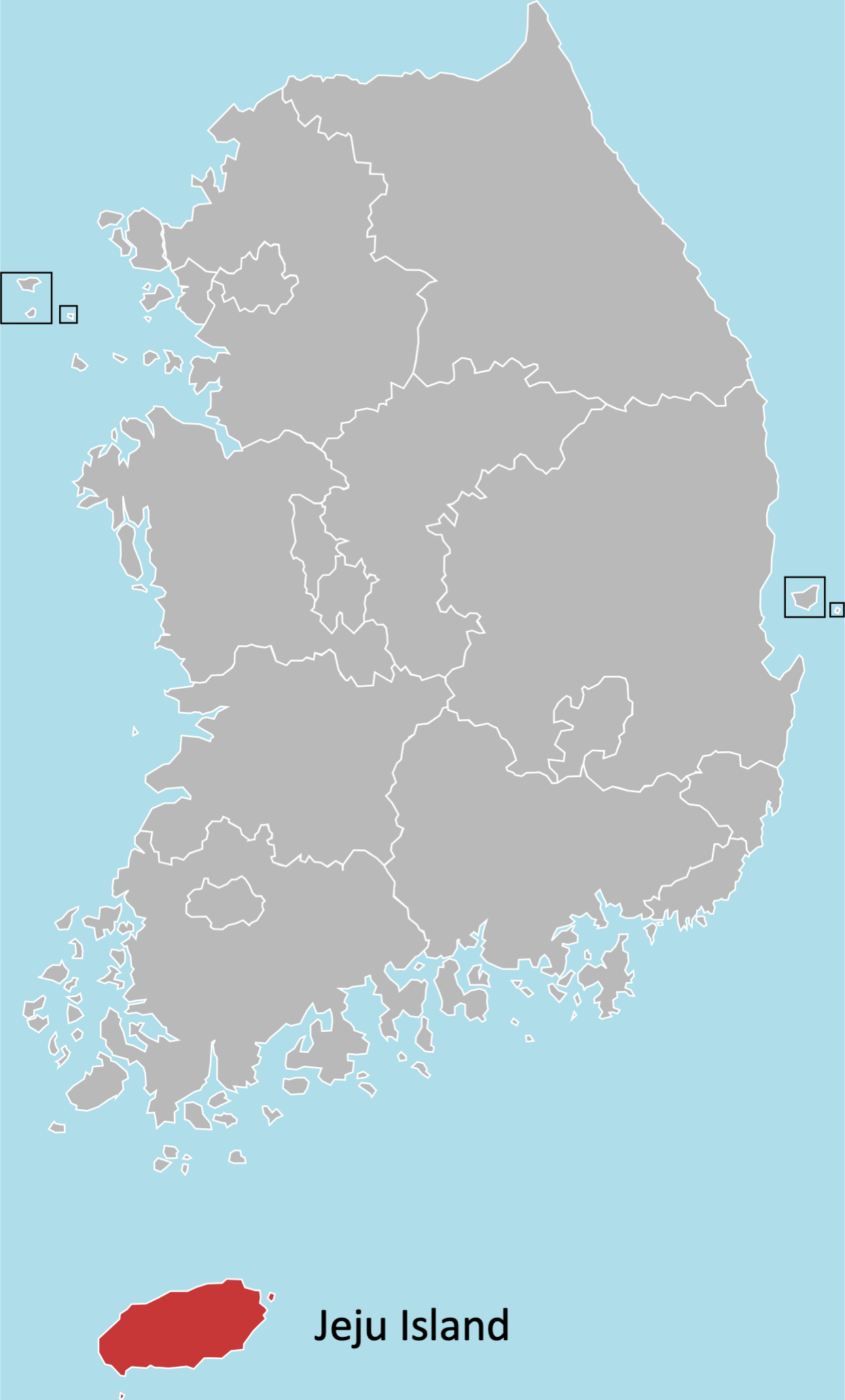}
    \caption{Jeju Island and South Korea.\protect\footnotemark}
    \label{fig:jeju}
\end{figure}

Jejueo (ISO 639-3 language code: jje) is the traditional language used on Jeju Island, located south of the Korean mainland (See Figure \ref{fig:jeju}). Today, there are only 5,000-10,000 fluent speakers, mostly above 70 years of age. The younger generation in Jeju is not learning the language in school, so they show a variety of levels of proficiency. 

For a long time, Jejueo has been treated as a dialect of Korean (ISO 639-3 language code: kor) rather than a distinct language \cite{o2015jejueo}. 
In this paper, we do not want to get into the debate of whether to consider Jejueo as a language or a dialect of Korean. Instead, we pay attention to the fact that Jejueo is often incomprehensible to Korean-only speakers \cite{yangrevising}. 
This motivates the need to consider Jejueo-Korean translation as an important language task.
In addition, Jejueo accent is different from standard Korean\footnote{Throughout this paper, Korean means standard Korean.}. So it sounds unnatural when a Korean speaker reads out a Jejueo text. In our preliminary experiment, we found out that our internal speech synthesis model for Korean was not able to generate Jejueo speeches properly.

For further information about Jejueo, we refer readers to the website of the Jejueo Project in University of Hawaii\footnote{\url{https://sites.google.com/a/hawaii.edu/jejueo}}. Here, we highlight one major difference between Jejueo and Korean: Araea (\begin{CJK}{UTF8}{mj}ㆍ\end{CJK}). Araea is a mid or low vowel that was used in Middle Korean. It is obsolete in contemporary Korean, but retained in Jejueo. Due to the presence of Araea, although both Jejueo and Korean are written in Hangul, Jejueo text is not easy for Korean speakers to type in digital settings. This will be discussed further in the next section. 

\footnotetext{\url{https://en.wikipedia.org/wiki/Jeju_language}}

\section{JIT (\underline{J}ejueo \underline{I}nterview \underline{T}ranscripts) Dataset} \label{jit}
The \textit{Jejueo Interview Transcripts} dataset, or JIT, is Jejueo-Korean parallel data compiled from \begin{CJK}{UTF8}{mj}
제주어구술자료집 1-20\end{CJK} by us. \begin{CJK}{UTF8}{mj}
제주어구술자료집\end{CJK}
 is the final report of the project performed by Center for Jeju Studies from 2014 until 2018. For the first three years, they interviewed Jeju senior citizens in Jejueo. Afterwards the interviews were carefully transcribed and then translated into standard Korean by experts. Along with additional notes, the results were arranged in 20 \texttt{pdf} files and opened to the public via their webpage\footnote{\url{http://www.jst.re.kr/}}. 
 
\begin{CJK}{UTF8}{mj}
제주어구술자료집\end{CJK} is an invaluable Jejueo resource in that it is arguably the largest Jejueo corpus publicly available. Unfortunately, it is not designed for computational use, after all. The \texttt{pdf} format is not machine friendly so it is tricky for researchers to work with it. Therefore, we convert the original \texttt{pdf} files into plain text files step by step so that they can be used for machine translation or any other computational tasks.

\begin{enumerate}
\item Convert \texttt{.pdf} files into plain \texttt{.txt} files using an online file conversion tool\footnote{\url{https://www.zamzar.com/convert/pdf-to-txt/}}.
\item Remove the front and back matters containing metadata. Accordingly, only interview dialogues remain.
\item Parse every line to extract Jejueo text and its Korean translation.
\begin{enumerate}
    \item In each line, Jejueo text is followed by Korean translation enclosed by parentheses. We capture Jejueo and Korean texts separately using simple regular expressions. However, some lines do not conform to the rule. For simplicity, we ignore those irregularities.
    \item Accidental line breaks frequently occur. We replace the line break with a special symbol, \^{}. It can take place in the middle of a word or between words. For example, \begin{CJK}{UTF8}{mj}제주도 날씨\end{CJK} `weather in Jeju Island' can have such forms as \begin{CJK}{UTF8}{mj}제\end{CJK}\^{}\begin{CJK}{UTF8}{mj}주도 날씨\end{CJK} or \begin{CJK}{UTF8}{mj}제주도\end{CJK}\^{}\begin{CJK}{UTF8}{mj}날씨\end{CJK}.
    \item Construct joint vocabulary, or a set of words. The words containing \^{} are removed from the vocabulary as they are yet incomplete.
    \item Check the incomplete words one by one and determine their real form. If the word without the \^{} is present in the vocabulary, the \^{} is removed. Otherwise, the \^{} is replaced by space. In the above examples, \begin{CJK}{UTF8}{mj}제\end{CJK}\^{}\begin{CJK}{UTF8}{mj}주도 날씨\end{CJK} becomes \begin{CJK}{UTF8}{mj}제주도 날씨\end{CJK} as \begin{CJK}{UTF8}{mj}제주도\end{CJK} is highly likely to appear somewhere else in the text, while \begin{CJK}{UTF8}{mj}제주도\end{CJK}\^{}\begin{CJK}{UTF8}{mj}날씨\end{CJK} becomes \begin{CJK}{UTF8}{mj}제주도 날씨\end{CJK} as \begin{CJK}{UTF8}{mj}제주도날씨\end{CJK} is not a (correct) single word.
  \end{enumerate}
\item Split punctuation marks into separate tokens.
\item Change private-use unicode characters into standard ones. Original text makes use of private-use areas in unicode to represent Araea (\begin{CJK}{UTF8}{mj}ㆍ\end{CJK}), a letter not used in contemporary Korean any longer. Not only can it cause unexpected issues but it is also against the unicode standard.
\item Shuffle and split the data into train, dev, and test sets. To avoid samples that are too short, the dev set and the test set are to have sentences of five words or more.
\end{enumerate}

As a result, we have 160,356, 5,000, and 5,000 Jejueo-Korean sentence pairs for train, dev, and test, respectively, as summarized in Table \ref{tbl:jit}. One thing to note is that the number of word forms in Jejueo are much larger than that in Korean (161,200 $>$ 110,774) although the total number of words in them is almost equal (1.4m). This is likely related to the fact that Jejueo speakers frequently use Korean as well as Jejueo, while Korean speakers do not. This will be further discussed in Section \ref{translation}

\setlength\tabcolsep{3pt}
\begin{table}[t]
    \centering
    \begin{tabular}{ccccc}
        \Xhline{1.1pt}
        \textbf{} & \textbf{Total} & \textbf{Train} & \textbf{Dev} & \textbf{Test} \\ \hline
        \# sentences & 170,356 & 160,356 & 5,000 & 5,000  \\  
        \# jje words & 1,421,723 & 1,298,672 & 61,448 & 61,603  \\  
        \# kor words & 1,423,836 & 1,300,489 & 61,541 & 61,806  \\ 
        \# jje word forms & 161,200  & 151,699 & 17,828& 18,029 \\
        \# kor word forms & 110,774 & 104,874 & 14,362 & 14,595 \\
        \Xhline{1.1pt}
    \end{tabular}
 
    \caption{Statistics of JIT dataset.}
    \label{tbl:jit}
\end{table}

\begin{figure}
    \centering
    %\caption{Schematic graphical model for words.}
    \includegraphics[height=5.5cm,clip=false]{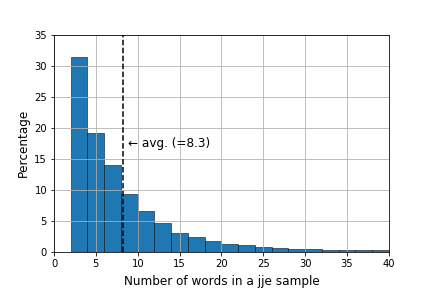}
    \caption{Number of words in a Jejueo sample of JIT dataset.}
    \label{fig:number_words}
\end{figure}

The length of Jejueo sentences ranges from 1 to 770 words. As can be seen in Figure \ref{fig:number_words}, however, most of them are 15 words or less. The average length is 8.3 words. Korean sentences show similar statistics.

\section{JSS (\underline{J}ejueo Single \underline{S}peaker \underline{S}peech) Dataset}
\subsection{Script}
We take the Jejueo text in the JIT dataset as the script for our speech dataset, \textit{Jejueo Single Speaker Speech} dataset (JSS). First, we randomly extract 10,000+ Jejueo sentences from the JIT dataset. To make the dataset more amenable to the training of speech synthesis models, we filter out ones which have more than 35 words or less than 3 words. Then the sentences that include any characters except space, Hangul, and punctuation marks are excluded as well. The final 10,000 sentences with their length information are written to a file in the tab separated format (\texttt{tsv}). As in Table \ref{tbl:jss}, the final 10k sentences are 9.4 words long on average. They amount to 94k words, or 335k characters.

\subsection{Audio}
We have an amateur voice actor record the script. He, in his thirties, was born in a rural area in Jeju and lived there until he was twenty. Although currently he does not stay in Jeju, he regularly visits his family back in Jeju, and speaks with them in Jejueo. He is instructed to read the script line by line as clearly and naturally as possible. Each sentence is saved as a \texttt{wav} file sampled at 44100 Hz. He works at his own pace for two months using his home recording devices. We trim the leading and trailing silence in the audio files using librosa\footnote{\url{https://librosa.github.io/librosa/}}. Finally, audio length is added to every line of the script. 

\begin{figure}
    \centering
    \includegraphics[height=5.2cm,clip=false]{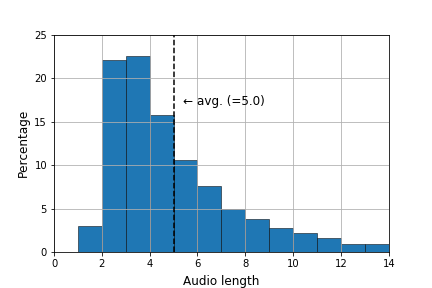}
    \caption{Durations of audio clips in JSS dataset. Most are between 2 and 8 seconds long. The average length of an audio clip is 5 seconds. }
    \label{fig:audio_len}
\end{figure}

The audio files are 13 hours and 47 minutes in total duration (Table \ref{tbl:jss}). Figure \ref{fig:audio_len} shows the distribution of the audio length. The shortest and the longest audio clips are 1.1 and 18.4 seconds long, respectively. Most of them, approximately 80\%, are 2 to 8 seconds long. The average length of an audio clip is 5 seconds.

\setlength\tabcolsep{3pt}
\begin{table}
    \centering
    \begin{tabular}{ccccc}
        \Xhline{1.1pt}
        \textbf{} & \textbf{Total} & \textbf{Avg.} & \textbf{Min.} & \textbf{Max.}  \\ \hline
        \# samples & 10,000 & - & - & -  \\ 
        \# words & 94,415 & 9.4 & 3 & 35  \\ 
        \# characters & 335,739 & 33.6 & 15 & 105  \\ % syl
        audio length & 13h 47m & 5.0s & 1.1s & 18.4s  \\ 
        
        \Xhline{1.1pt}
    \end{tabular}
 
    \caption{Statistics of JSS dataset}
    \label{tbl:jss}
\end{table}

\section{Jejueo-Korean Machine Translation} \label{translation}
Using the JIT dataset, we train machine translation models between Jejueo and Korean.
We consider translation in both directions, kor $\rightarrow$ jje and jje $\rightarrow$ kor, and evaluate the performance of each model by computing the BLEU scores \cite{papineni2002bleu} on the dev/test set.

\subsection{Model \& Setup}
Throughout our experiments, we use the Transformer \cite{vaswani2017attention}, a state-of-the-art model for neural machine translation.
The Transformer is a deep sequence-to-sequence (seq2seq) architecture primarily based on attention mechanisms, including both an encoder-decoder attention \cite{bahdanau2015neural,luong2015effective} and self-attention \cite{lin2017structured}. 

We follow the original parameter settings of the standard Transformer model: 6 encoder and decoder blocks, each with 512-2048 hidden units across 8 attention heads.
We run all of our experiments using \textsc{fairseq}\footnote{\url{https://github.com/pytorch/fairseq}} \cite{ott2019fairseq}, a PyTorch-based library for deep sequence models.
Details of the training procedure, including all hyperparameters, can be found in our GitHub repository.

\subsection{Choosing Optimal Vocabulary Size}\label{sec:vocab}

Byte Pair Encoding (BPE) is a simple data compression technique that iteratively replaces  the  most  frequent  pair  of  bytes  in text with a single, unused byte \cite{Gage:1994:NAD:177910.177914}. Since \cite{sennrich2016neural} successfully applied it to neural machine translation models, it has been a \emph{de facto} standard in the word segmentation for machine translation. Therefore, we also apply BPE to our models.

We first run an experiment to determine the optimal BPE vocabulary size for Jejueo-Korean translation.
For various vocabulary size options, we tokenise input text using SentencePiece\footnote{\url{https://github.com/google/sentencepiece}} \cite{kudo2018sentencepiece}.
The vocabulary is shared between the encoder and the decoder. 

In Table \ref{tbl:vocab_size}, we summarize our results using five vocabulary sizes: 2k, 4k, 8k, 16k, and 32k. 
We find that using 4k vocabulary size leads to the best BLEU scores on the dev/test set for both kor $\rightarrow$ jje (44.85/43.31) and jje $\rightarrow$ kor (69.35/67.70), although they are within a point difference for 2k and 8k vocabulary sizes. 
Performance degrades for using larger vocabulary sizes: by approximately 1 point for 16k and another 1 point for 32k.

\subsection{Comparison with Copy Models}

Using the Transformer model with 4k vocabulary size, we present our main baselines in Table \ref{tbl:copy}.
As a simple baseline, we include a copying model that predicts its input as its output (``Copy'').
The copying model already achieves 24.44 and 24.45 BLEU scores on the kor $\rightarrow$ jje and jje $\rightarrow$ kor test sets respectively.
By training a Transformer model on the JIT dataset (``JIT''), the scores significantly improve to 43.31 and 67.70 respectively, as we illustrated in Section \ref{sec:vocab}

We remark that the BLEU scores of the jje $\rightarrow$ kor models (65-67) are much higher than those of the kor $\rightarrow$ jje models (41-43). One possible explanation is that Korean as well as Jejueo is frequently used in Jejueo dialogues as we discussed in Section \ref{jejueo} For example, when \begin{CJK}{UTF8}{mj}아버지\end{CJK} `father' appears in the Korean text of the JIT dataset, \begin{CJK}{UTF8}{mj}아버지\end{CJK} co-occurs 530 times in the paired Jejueo text, while the Jejueo equivalent, \begin{CJK}{UTF8}{mj}아방\end{CJK}, does only 332 times. In short, that a Korean word can correspond to either a Jejueo counterpart or itself makes the kor $\rightarrow$ jje translation harder than the other direction. 

The fact that copying models without training achieve non-trivial BLEU scores implies that the JIT model may benefit from additional training on a copying task.
To test this idea, we follow the approach taken by \cite{sennrich-etal-2016-improving} and augment both the source and target sides of the training set with the same number of randomly sampled Korean sentences from a Wikidump\footnote{\url{https://dumps.wikimedia.org/kowiki/20190601/}} (``JIT + KorWiki'').
This further improves the dev/test set BLUE scores by up to 0.88 points: 44.19 for kor $\rightarrow$ jje and 67.94 for jje $\rightarrow$ kor.

\begin{table}
    \centering
    \begin{tabular}{cccc}
        \Xhline{1.1pt}
        \textbf{Lang. Pair} & \textbf{\# Vocab.} & \textbf{Dev} & \textbf{Test} \\ \hline
        kor $\rightarrow$ jje 
         & 2k &  44.80 & 43.26 \\ % 200:60
         & \textbf{4k} &  \textbf{44.85} & \textbf{43.31} \\ % 600:55
         & 8k   & 44.40 & 43.03 \\ % 500:65
         & 16k & 43.33  & 42.08 \\ % 400:70
         & 32k & 42.57  & 41.07 \\ \hline % 300:55 
         jje $\rightarrow$ kor 
         & 2k &  69.05 & 67.63 \\ % 1000:65
         & \textbf{4k} & \textbf{69.35} & \textbf{67.70} \\ % 600:70
         & 8k & 69.02 &  67.46 \\ % 800:75
         & 16k  & 67.61 & 66.30 \\ % 900:95
         & 32k  & 66.32 & 65.08 \\ % 900:100
        \Xhline{1.1pt}
    \end{tabular}
 
    \caption{BLEU scores of models according to the different BPE vocabulary size. SentencePiece is used for BPE segmentation. All hyperparameters except the vocabulary size are identical.}
    \label{tbl:vocab_size}
\end{table}

\begin{table}
    \centering
    \begin{tabular}{cccc}
        \Xhline{1.1pt}
        \textbf{Lang. Pair} & \textbf{Model} & \textbf{Dev} & \textbf{Test} \\ \hline
        kor $\rightarrow$ jje & Copy & 24.06 & 24.44  \\ 
         & JIT & 44.85 & 43.31  \\ %4k
         & JIT + KorWiki & \textbf{45.25}  & \textbf{44.19} \\ \hline % 33 
         jje $\rightarrow$ kor & Copy &  24.07 & 24.45 \\ 
         & JIT & 69.35 & 67.70  \\ %4k
         & JIT +  KorWiki & \textbf{69.59} & \textbf{67.94}  \\ % 66 
        \Xhline{1.1pt}
    \end{tabular}
    \caption{BLEU scores of various translation models. In the Copy model, translation outputs are copied from the source. For the JIT + KorWiki model, 160,356 Korean sentences extracted from a Wikidump are added to both source and target sides of the JIT dataset. The vocabulary size is fixed to 4k.}
    \label{tbl:copy}
\end{table}

\begin{table*}[!ht]
    \centering
    \begin{tabular}{ccccccc}
        \Xhline{1.1pt}
        \textbf{Token Type} & \textbf{Unicode Range} & \textbf{\# Vocab.} & \textbf{Length} & \textbf{Example 1} & \textbf{Example 2} & \textbf{MCD} (mean / std.)\\ \hline
        character &   U+AC00-D7AF & 1,412 & 34 & \begin{CJK}{UTF8}{mj}국\end{CJK} & \begin{CJK}{UTF8}{mj}쉐똥\end{CJK} & 14.47/0.59  \\ 
        Hangul Jamo & U+1100-11FF & 74  & 64 & \begin{CJK}{UTF8}{mj}ㄱ\textsuperscript{onset}ㅜㄱ\textsuperscript{coda}\end{CJK} & \begin{CJK}{UTF8}{mj}ㅅㅞㄸㅗㅇ\end{CJK} & \textbf{14.32/0.38} \\ 
        Hangul Jamo (S) & U+1100-11FF & 59 & 65 & \begin{CJK}{UTF8}{mj}ㄱ\textsuperscript{onset}ㅜㄱ\textsuperscript{coda}\end{CJK} & \begin{CJK}{UTF8}{mj}ㅅㅞㄷㄷㅗㅇ\end{CJK} & 14.34/0.43 \\
        HCJ & U+3130-318F & 57 & 64 & \begin{CJK}{UTF8}{mj}ㄱㅜㄱ\end{CJK} & \begin{CJK}{UTF8}{mj}ㅅㅞㄸㅗㅇ\end{CJK} &  14.46/0.62 \\
        HCJ (S) & U+3130-318F & 44 & 65 & \begin{CJK}{UTF8}{mj}ㄱㅜㄱ\end{CJK} & \begin{CJK}{UTF8}{mj}ㅅㅞㄷㄷㅗㅇ\end{CJK} & 14.48/0.44 \\
        \Xhline{1.1pt}
    \end{tabular}
    \caption{Mel Cepstrum Distortion values on the 100 test samples. (S) denotes single consonants only. Note that although the examples of Hangul Jamo and HCJ may look the same, actually they are different in code point. 
    The MCD values of Jamo are the lowest. The lower, the better.}
    \label{tbl:speech_models}
\end{table*}

\section{Jejueo Speech Synthesis}
\subsection{Model}
We  train  a  Jejueo Text-To-Speech (TTS) model called DCTTS \cite{tachibana2018dctts}, on the JSS dataset. For the past years there have been many neural TTS models such as WaveNet \cite{oord2016wavenet}, Tacotron 1 \& 2 \cite{wang2017tacotron,shen2018natural}, Char2Wav \cite{sotelo2017char2wav}, DeepVoice 1-3 \cite{arik2017deep,gibiansky2017deep,ping2018deep}, and VoiceLoop \cite{taigman2017voiceloop}. Among them, DCTTS is lightweight and fast because it is made up of convolution layers only. Besides, thanks to several tricks such as guided attention and incremental attention, its training is stable. With our working implementation which were already successfully used in \cite{park2019css10}, we conduct experiments on the JSS dataset.

For training each model, we mostly adopt the hyperparameters in \cite{tachibana2018dctts}. 
Compared to the original implementation, we additionally add dropout \cite{srivastava2014dropout} of 0.05 to every layer for regularization. 
We train all models for 200k steps. Among the 10k JSS samples, the last 100 samples are held out for test. 

\subsection{Finding the Best Token Type}
In most neural TTS systems, either graphemes (spelling) or phonemes (pronunciation) are taken as input. For a script that is not phonetic, e.g., Chinese characters, grapheme-to-phoneme conversion is considered compulsory. However, as Hangul is phonetic, in other words, text in Hangul sounds as it is written, we stick with graphemes rather than converting them into phonemes.

Throughout our experiments, we examine which token unit works the best for Jejueo speech synthesis.
In truth, a Hangul character is a syllable, and can be decomposed into its constituent vowels and consonants. They are called \textit{Jamo} in Korean. This strategy is helpful for readability in practice, but brings about the following question: do we have to break Hangul syllables into \textit{Jamo} in Jejueo speech synthesis? 

\textit{Jamo} has two character blocks in unicode: Hangul Jamo (U+1100-11FF) and Hangul Compatibility Jamo (HCJ) (U+3130-318F).
Their major difference is that, in HCJ, syllable-initial consonants (onset) are reused as syllable-final consonants (coda), whereas in Hangul Jamo onset and coda are two separate sets.
In Example 1 of Table \ref{tbl:speech_models}, the character \korean{국} is decomposed into a vowel (\korean{ㅜ}) and consonants (both \korean{ㄱ}) in the Hangul Jamo and HCJ rows.
Note that in Hangul Jamo, the onset \korean{ㄱ} and the coda \korean{ㄱ} are treated as separate characters unlike in HCJ.
A further distinction can be made according to whether or not we break consonant clusters in \textit{Jamo} such as  \begin{CJK}{UTF8}{mj}ㄲ, ㄸ\end{CJK}, or \begin{CJK}{UTF8}{mj}ㄵ\end{CJK} into a sequence of letters, i.e., \begin{CJK}{UTF8}{mj}ㄱㄱ, ㄷㄷ\end{CJK}, and \begin{CJK}{UTF8}{mj}ㄴㅈ\end{CJK}. In Example 2 of Table \ref{tbl:speech_models}, the \begin{CJK}{UTF8}{mj}ㄸ\end{CJK} in Hangul Jamo and HCJ is segmented into \begin{CJK}{UTF8}{mj}ㄷㄷ\end{CJK} in their (S) versions.

In the first five columns of Table \ref{tbl:speech_models}, we summarize various tokenisation strategies we compare in our experiments.

\subsection{Evaluation \& Results}
TTS systems are commonly evaluated with Mean Opinion Score (MOS), the arithmetic mean over all values in the range 1-5 given by individuals. Although the MOS is widely used, it is inherently weak to biases as it is subjective. Besides, it is costly so scalablity is low. For these reasons, we evaluate the performance of each model using a Mel Cepstral Distortion (MCD) measure in this study. It is the average Euclidean distance between the mel cepstral feature vectors of reference and synthesized audio files. So generally speaking, the lower the MCD value is, the better the audio quality is.

For each model, we synthesize 100 audio samples based on the last 100 lines of the script that were not used for training. As shown in Table \ref{tbl:speech_models}, we find that the MCD mean values of the Hangul Jamo model are the lowest of all. In other words, the audios synthesized by the Hangul Jamo model are most similar to the original ones. We believe it is because the Hangul Jamo model has more granular information than all the others. In terms of granularity, the Hangul Jamo model performs better than the character model, possibly because in the latter vowels and consonants are hidden in the syllable. It also outperforms the HCJ models, as the former has two different sets for a consonant, unlike the latter. Finally, the Hangul Jamo model has more information than its single consonants only version, which replaces consonant clusters with a sequence of single consonants.

\section{Conclusion}

In this paper, we presented two new Jejueo datasets, JIT and JSS, and explained why and how we developed them.
The JIT dataset is bilingual data where 170k+ Jejueo sentences are paired with their Korean translations. 
The JSS dataset consists of 10k high-quality audio files recorded by a Jejueo speaker and a transcript file. We carried out two follow-up tasks: Jejueo-Korean machine translation and Jejueo speech synthesis using those datasets. 
In our experiments, neural machine translation models of 4k shared BPE vocabulary and a neural speech synthesis model based on Hangul Jamo tokens showed the best performance. 
We hope that our datasets will attract a lot of attention from both language and machine learning communities. 

\section{Acknowledgements}

We express our deep respect to Center for Jeju Studies for their devotion to \begin{CJK}{UTF8}{mj}제주어구술자료집\end{CJK} where this project began. We also thank Soo Kyung Lee of Kakao Brain for her support.

\section{References}

\bibliographystyle{lrec}
\bibliography{lrec2020W-xample}

\end{document}